\documentclass[letterpaper]{article}
\usepackage{aaai}
\usepackage{times}
\usepackage{helvet}
\usepackage{courier}
\usepackage[frozencache,cachedir=.]{minted}

\begin{document}
\title{srlearn: A Python Library for Gradient-Boosted Statistical Relational Models}
\author{Alexander L. Hayes\\
ProHealth Lab\\
Indiana University Bloomington\\
hayesall@iu.edu
}
\maketitle
\begin{abstract}
\begin{quote}
We present \texttt{srlearn}, a Python library for boosted statistical relational models.
We adapt the scikit-learn interface to this setting and
provide examples for how this can be used to express learning and inference
problems.
\end{quote}
\end{abstract}

\section{Introduction}

Traditional machine learning systems have generally been built as command line
applications or as graphical user interfaces \cite{hall2009weka}. Both have advantages,
but offer fewer solutions when data acquisition, preprocessing, and model
development must occur together. Systems such as scikit-learn,
TensorFlow, Pyro, and PyTorch solve
this problem by embedding data cleaning and model development as steps
within general-purpose languages
\cite{pedregosa2011scikit,abadi2016tensorflow,bingham2019pyro,paszke2017pytorch}.
This has also made open source implementations available to both experts
and non-experts, providing each the tools to develop models.

Statistical Relational Learning (SRL) models have unique concerns, often
inherited from underlying logical systems. This requires a
data representation beyond fixed-length feature vectors, and a language
bias to constrain the hypothesis space. By embedding both
operations in a manner that machine learning researchers and practitioners
may already be familiar with, we hope to speed up development time for SRL practitioners,
and provide a more user-friendly experience for data scientists and the wider machine
learning community---many of whom are not experts in SRL.

\subsection{API Design in Machine Learning}

The scikit-learn package \cite{pedregosa2011scikit} has been influential
for its consistent application programming interface (API) across a variety of machine learning models.
In scikit-learn, an algorithm type (e.g. linear support vector classification) is implemented as a class.
An \textit{estimator} is an instance of an algorithm type whose hyperparameters have been set upon object construction.
A \textit{predictor} is an estimator that has been \texttt{fit} (i.e. trained) to a dataset, and is ready to \texttt{predict} (e.g. classify) new data instances.
Although the estimation and prediction functions are logically distinguished in two separate protocols, it is generally a single class that implements both a learning algorithm and the model for applying the parameters to new data.

The aforementioned standard approach thus comprises configuring model hyperparameters, fitting training data, and predicting test data.
However, scikit-learn has since developed into a full-fledged ecosystem that also services related functions in the modeling workflow: model selection, hyperparameter tuning, and model validation.
Furthermore, multiple offshoots of scikit-learn have emerged to tackle more specialized challenges, including imbalanced datasets \cite{lemaitre2017imbalancedlearn},  generalized linear models \cite{blondel2016lightning}, and metric learning \cite{vazelhes2019metric}.

These offshoots still fit within the framework of only requiring inputs, outputs, and hyperparameters. But while this has been influential while designing APIs for classic statistical learning methods, it could also be a limitation when extending the API to incorporate the specific needs of models from other learning paradigms \cite{buitinck2013apidesign}.
Learning within frameworks designed for graphical models, active learning, or reinforcement learning typically requires the user to specify something outside of inputs and outputs.
Graphical models require statistical independence assumptions to either be set by hand or inferred via structure learning.
Active learning requires human intervention.
Reinforcement learning needs a simulator.

Extending the API to handle new paradigms should ideally meet two goals:
(1) \textit{expressiveness} to describe what the user wants to achieve, and
(2) \textit{complimentarity} to what users are already familiar with.

\section{\texttt{srlearn}}

We propose a simple extension to the scikit-learn API for representing
statistical relational models while staying close to our two goals.
Specifically we incorporate a \texttt{Background} object and a
\texttt{Database} object.

The \texttt{Background} object incorporates knowledge about relationships
to constrain model search space, currently expressed in the language
of ``modes'' \cite{srinivasan2004aleph}. This is then provided
to the statistical relational estimator.

The \texttt{Database} object generalizes inputs as being composed of
positive examples, negative examples, and facts about the world---each
expressed as Prolog predicates.

A statistical relational estimator may then be described in the same
language as a standard scikit-learn estimator that also incorporates
background knowledge to constrain the hypothesis space, and learn
on a database of predicates rather than vectors. Currently
we have focused on incorporating methods from \textsc{BoostSRL}, a Java tool
for learning relational dependency networks and Markov
logic networks via gradient boosting \cite{natarajan2018human}.
Figure~\ref{minted:smokes_friends_cancer} shows how modules from \texttt{srlearn}
can be put together to learn on a built-in data set, then make
predictions on a test database.

\begin{figure}
\begin{minted}[fontsize=\footnotesize]{python}
from srlearn.rdn import BoostedRDN
from srlearn import Background
from srlearn import example_data

bk = Background(
  modes=[
    "friends(+person,-person).",
    "friends(-person,+person).",
    "cancer(+person).",
    "smokes(+person).",
  ]
  use_std_logic_variables=True,
)

clf = BoostedRDN(
  background=bk,
  target="cancer",
)

clf.fit(example_data.train)
clf.predict_proba(example_data.test)
\end{minted}
\caption{
Learning and inference on toy databases for a smokes-friends-cancer domain.
\texttt{example\_data.train} and \texttt{example\_data.test} are \texttt{Database} objects.
}
\label{minted:smokes_friends_cancer}
\end{figure}

\subsection{Development}

\texttt{srlearn} is developed as an open source project on
GitHub\footnote{https://github.com/hayesall/srlearn/} and is distributed
under the terms of the GNU General Public License v3.0 (GPL-3.0).
Within the code, we have taken several measures to aid its
maintenance. This includes formatting conventions (\texttt{black}, \texttt{pycodestyle}),
linting (\texttt{pylint}), and running the main branch and all pull
requests through static analysis (\texttt{lgtm}).

We also maintain a test suite to compare each
build against previous versions. Tests run on Linux and Windows
machines each time the code is pushed to GitHub; metrics track
(1) that all tests pass, and (2) that a
sufficient code coverage is maintained. At the time of writing,
all tests pass (results meet expectations), and code coverage
is at 100\% (every line of code is visited during testing).
Perfect coverage often grows unrealistic as projects
grow, but we aim to keep it above 90\% while passing all tests.

Finally, we maintain documentation\footnote{https://srlearn.readthedocs.io}
to help acclimate users to the
code base; this includes user guides with narrative documentation and
examples motivating specific tasks.

\subsection{Experiments}

\begin{table}
    \centering
    \begin{tabular}{l|ll|}
     & \multicolumn{1}{p{3cm}}{\texttt{srlearn}\newline (Python/Java)}
     & \multicolumn{1}{p{1.8cm}|}{\textsc{BoostSRL}\newline (Java/Shell)} \\
     \hline
    WebKB & 4.2 (0.5) & 4.9 (0.3) \\
    IMDB & 10.2 (1.1) & 13.0 (1.3) \\
    UWCSE & 17.5 (1.4) & 18.3 (1.7) \\
    \end{tabular}
    \caption{
        Seconds elapsed while learning a Boosted RDN on three benchmark data sets.
        Mean (and standard deviation) are calculated over ten runs. Small differences in
        times may also be influenced by small differences in measurement: epoch time (Bash)
        and \texttt{perf\_time} (Python).
    }
    \label{table:experiments}
\end{table}

We expect a small overhead due to the Python interpreter and data
structures at runtime; but since the core algorithms borrow heavily \textsc{BoostSRL}'s
Java implementations, we expect this overhead to be negligible compared to the time
spent during learning.
To evaluate this, we compare runtime in seconds on standard benchmark data using
the \textsc{BoostSRL} command line interface and the \texttt{srlearn} API.
We hold the modes and hyperparameters fixed,
then record the time taken while learning a boosted RDN with the \texttt{srlearn} and
\textsc{BoostSRL} systems on three benchmark data sets. Table~\ref{table:experiments}
shows the time averaged over ten runs of each, which we use to conclude
that the time differences are indeed
negligible.\footnote{Scripts for reproducing this table is available on GitHub: https://github.com/hayesall/srlearn-StarAI-2020-workshop}

\section{Conclusion}

It is possible that the imperative programming style here is not ideal
for SRL models---the underlying logic formalism is often better
expressed through declarative approaches, which have further been suggested as ways to unify software
development with learning systems \cite{kordjamshidi2018systems}.

Nonetheless, many learning frameworks have been built around the
Python ecosystem. Programming abstractions such as the one presented here
may therefore be an important step toward bridging the gap between SRL
and neural approaches by providing developers the tools to more easily work with both in
a common environment.

In the future, we intend on extending the modeling language with more methods
that have been successful within SRL---such as learning
with advice \cite{odom2018human}, incorporating a relational database for learning
and inference \cite{malec2017inductive}, and incorporating SRL methods such as
Probabilistic Soft Logic \cite{bach2015hinge} or Conditional Random Fields \cite{sutton2007inductive}.

\section{Acknowledgements}

ALH is supported through Indiana University's ``Precision Health Initiative" (PHI) Grand Challenge.
ALH would like to thank Sriraam Natarajan, Travis LaGrone, and members of the StARLinG Lab at the University of
Texas at Dallas.

\bibliographystyle{aaai}
\bibliography{hayesall_srlearn}

\end{document}